\documentclass{article} % For LaTeX2e
\usepackage{iclr2025_conference,times}

\usepackage{amsmath,amsfonts,bm}

\def\eqref#1{equation~\ref{#1}}
\def\1{\bm{1}}

\DeclareMathAlphabet{\mathsfit}{\encodingdefault}{\sfdefault}{m}{sl}
\SetMathAlphabet{\mathsfit}{bold}{\encodingdefault}{\sfdefault}{bx}{n}

\usepackage{hyperref}
\usepackage{url}
\usepackage{caption}

\usepackage{latexsym}

\usepackage{graphicx}
\usepackage{multirow}
\usepackage{array}
\usepackage{booktabs}
\usepackage[table]{xcolor}
\definecolor{lightgray}{gray}{0.9}
\usepackage{enumitem}
\usepackage{url}
\usepackage{wrapfig}

\usepackage{amssymb}        % 用于 \checkmark 符号
\usepackage{amsmath}        % 用于 \boldsymbol 加粗数学符号
\usepackage{makecell}       % 用于创建多行单元格

\usepackage{pifont}         % 用于生成更粗的勾和叉符号
\usepackage{microtype}

\usepackage{siunitx}
\usepackage{tabularx}

\usepackage[most]{tcolorbox}
\newtcolorbox{promptbox}{
    colback=white,
    colframe=black,
    boxrule=1pt,
    rounded corners,
    boxsep=0.5mm, % 框内边距
    breakable,
}

\setlist[description]{
    style=nextline,    % 强制内容换到下一行
    leftmargin=0pt,    % 整个列表不缩进
    itemindent=1.5em,  % 内容行（第二行）的缩进量
    labelsep=0pt,      % 标签和内容之间的水平间距（此处不需要）
}

\usepackage[T1]{fontenc}
\usepackage[utf8]{inputenc}

\usepackage{microtype}

\usepackage{inconsolata}

\usepackage{graphicx}

\definecolor{mygreen}{HTML}{4CAF50}
\definecolor{myred}{HTML}{D9534F}

\newcommand{\cmark}{\textcolor{mygreen}{\ding{51}}}
\newcommand{\xmark}{\textcolor{myred}{\ding{55}}}
\title{CorpusQA: A 10 Million Token Benchmark for Corpus-Level Analysis and Reasoning}

\author{
 \textbf{Zhiyuan Lu\textsuperscript{1}},
 \textbf{Chenliang Li\textsuperscript{1}},
 \textbf{Yingcheng Shi\textsuperscript{1}},
 \textbf{Weizhou Shen\textsuperscript{1}},
 \textbf{Ming Yan\textsuperscript{1}},
 \textbf{Fei Huang\textsuperscript{1}}
\\
\\
 \textsuperscript{1}Tongyi Lab, Alibaba Group
\\
 \small{
    \textbf{Correspondence:} {zylu734@gmail.com, ym119608@alibaba-inc.com}
  }
}

\iclrfinalcopy % Uncomment for camera-ready version, but NOT for submission.
\makeatletter
\def\ps@plain{%
  \let\@oddhead\@empty
  \let\@evenhead\@empty
  \def\@oddfoot{\hfil\thepage\hfil}%
  \def\@evenfoot{\hfil\thepage\hfil}%
  \def\@oddhead{\vbox{\hbox to\textwidth{\hfil\thepage}%
    \vskip 1pt\hrule height 0.4pt}}%
  \let\@evenhead\@oddhead
}
\makeatother
\begin{document}
\maketitle

\begin{abstract}
While large language models now handle million-token contexts, their capacity for reasoning across entire document repositories remains largely untested. Existing benchmarks are inadequate, as they are mostly limited to single long texts or rely on a "sparse retrieval" assumption—that answers can be derived from a few relevant chunks. This assumption fails for true corpus-level analysis, where evidence is highly dispersed across hundreds of documents and answers require global integration, comparison, and statistical aggregation. To address this critical gap, we introduce CorpusQA, a new benchmark scaling up to 10 million tokens, generated via a novel data synthesis framework. By decoupling reasoning from textual representation, this framework creates complex, computation-intensive queries with programmatically guaranteed ground-truth answers, challenging systems to perform holistic reasoning over vast, unstructured text without relying on fallible human annotation. We further demonstrate the utility of our framework beyond evaluation, showing that fine-tuning on our synthesized data effectively enhances an LLM's general long-context reasoning capabilities. Extensive experiments reveal that even state-of-the-art long-context LLMs struggle as input length increases, and standard retrieval-augmented generation systems collapse entirely. Our findings indicate that memory-augmented agentic architectures offer a more robust alternative, suggesting a critical shift is needed from simply extending context windows to developing advanced architectures for global information synthesis.
\end{abstract}

\section{Introduction}
\begin{wrapfigure}{r}{0.5\columnwidth}
    \vspace{-10pt}
    \centering
    \includegraphics[width=0.5\columnwidth]{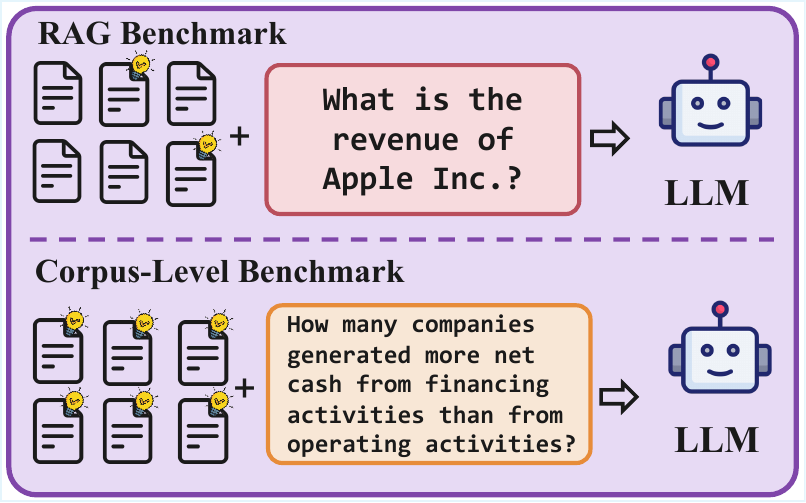}
    \caption{RAG-style benchmarks assume sparse retrieval, where the answer can be derived from a few retrieved documents. In contrast, corpus-level benchmarks require holistic reasoning over an entire document repository with highly dispersed evidence.}
    \vspace{-10pt}
    \label{fig:cover}
\end{wrapfigure}

Recent advancements in large language models have pushed the boundaries of context processing, with models now capable of ingesting millions of tokens in a single prompt\cite{comanici2025gemini,yang2025qwen2,chen2025minimax}. More recently, efforts have been made to extend the complex reasoning capabilities of reinforcement learning models to long-context scenarios~\cite{wan2025qwenlong}. This has unlocked remarkable capabilities in long-document understanding and summarization. However, a critical gap persists between processing a single long document and performing analysis across an entire document repository. In high-stakes professional domains such as finance, law, and scientific research, real-world tasks often demand reasoning not over one text, but over a collection of hundreds or thousands of documents to uncover insights that are only visible at a corpus level. For instance, a financial analyst might need to determine how many companies generated more net cash from financing activities than from operating activities after reviewing dozens of financial reports (Figure~\ref{fig:cover}), or a legal researcher might need to synthesize precedents from an entire case law database. We term this challenging paradigm Corpus-Level Analysis.

\begin{table*}[t]
\centering
\small
\renewcommand{\arraystretch}{0.9}
\setlength{\tabcolsep}{4pt} % 可选：减小列间距以更好适应

\begin{tabularx}{\textwidth}{l *{5}{>{\centering\arraybackslash}X}}
\toprule
\textbf{Benchmark} & 
\makecell{Up to 10M \\ Length} & 
\makecell{Reasoning \\ Intensive} & 
\makecell{Controllable \\ Context} & 
\makecell{High Evidence \\ Dispersion} & 
\makecell{Multilingual} \\
\midrule
\textbf{InfiniteBench}~\cite{zhang2024infty} & \xmark & \xmark & \xmark & \xmark & \cmark \\
\textbf{LongBench v2}~\cite{bai2024longbench}  & \xmark & \cmark & \xmark & \xmark & \cmark \\
\midrule
\textbf{Loong}~\cite{wang2024leave}         & \xmark & \xmark & \cmark & \cmark & \cmark  \\
\textbf{FanOutQA}~\cite{zhu2024fanoutqa}      & \xmark & \xmark & \cmark & \cmark & \xmark  \\
\midrule
\textbf{FRAMES}~\cite{krishna2024factfetchreasonunified}        & \xmark & \cmark & \xmark & \xmark & \xmark  \\
\textbf{CRAG}~\cite{yang2024crag}          & \xmark & \xmark & \xmark & \xmark & \xmark \\
\midrule
\textbf{CorpusQA (Ours)}& \cmark & \cmark & \cmark & \cmark & \cmark  \\
\bottomrule
\end{tabularx}
\caption{Comparison of CorpusQA with existing long-context and RAG benchmarks. CorpusQA is the first benchmark designed to simultaneously evaluate reasoning over massive-scale corpora (up to 10M tokens) with highly dispersed evidence, addressing a key gap in prior work.}
\label{tab:benchmark_comparison}
\vspace{-5pt}
\end{table*}

A defining characteristic of this paradigm is the high dispersion of evidence, where critical information is not concentrated in a few documents but is finely scattered across the entire repository. Consequently, deriving a correct answer demands a holistic processing of the complete, ultra-long context, coupled with complex reasoning, statistical aggregation, and precise computation. This requirement directly invalidates the core assumption of Retrieval-Augmented Generation (RAG)~\cite{lewis2020retrieval,zhao2024retrieval,yu2024evaluation}, which presupposes that answers can be synthesized from a small number of relevant retrieved chunks, rendering it inherently unsuitable for such globally-aware tasks.

% Existing evaluation benchmarks are ill-equipped to measure this capability. They fall into three main categories, each with fundamental limitations. (1) Long-context benchmarks like LongBenchV2~\cite{bai2024longbench} and InfiniteBench~\cite{zhang2024infty} primarily evaluate a model's ability to recall facts or reason within a single, concatenated text. They do not address the challenge of integrating information scattered across a multitude of discrete documents. (2) Multi-document QA benchmarks such as Loong~\cite{wang2024leave} and MDBench~\cite{peper2025mdbench} take a step in the right direction, but typically involve fewer than a dozen documents. The total context size often remains within the capacity of modern long-context LLMs, reducing the problem to a variant of single-document QA rather than a true corpus-level challenge. (3) RAG benchmarks like CRAG~\cite{yang2024crag} and FRAMES~\cite{krishna2024factfetchreasonunified} are predicated on a "sparse retrieval" assumption: that the answer can be synthesized from a few highly relevant retrieved chunks. This paradigm breaks down for corpus-level tasks, where evidence is densely dispersed and a global view of the entire repository is necessary for accurate statistical calculation, trend analysis, or comparative reasoning. Table \ref{tab:benchmark_comparison} summarizes these limitations and contrasts them with the capabilities of CorpusQA.

Existing evaluation benchmarks~\cite{krishna2024factfetchreasonunified,yang2024crag} are inadequate to measure this capability. They are largely confined to single long texts or rely on a "sparse retrieval" assumption that is invalidated by the high evidence dispersion in corpus-level analysis. Even recent multi-document benchmarks~\cite{wang2024leave,zhu2024fanoutqa} that require integrating information across all provided texts operate at a scale too small to pose a true corpus-level challenge, as their entire context often fits within a single prompt for modern LLMs. This leaves a critical gap in assessing holistic reasoning over vast repositories, a gap CorpusQA is designed to fill. Table \ref{tab:benchmark_comparison} summarizes these limitations and contrasts them with our approach.

To bridge this gap and steer research towards this critical challenge, we introduce CorpusQA, the first benchmark designed for large-scale corpus-level analysis. Our benchmark makes several key departures from prior work by combining three core design principles. It features \textbf{unprecedented scale and complexity}, scaling to 10M tokens with reasoning-intensive queries that necessitate architectures for global synthesis over simple context expansion. This scale enables our second principle: \textbf{high information dispersion}, where deliberately scattered evidence invalidates RAG's sparse retrieval assumption and necessitates a holistic understanding. Finally, to ensure reliable evaluation against this challenge, we introduce \textbf{guaranteed factual grounding}. We developed a novel schema-driven pipeline that programmatically computes ground-truth answers from an underlying structured representation of the documents, thus providing verifiable correctness without relying on a fallible LLM for annotation.

We conduct extensive experiments on CorpusQA, evaluating a wide array of state-of-the-art long-context LLMs, RAG systems, and more advanced agentic architectures. Our findings reveal a clear picture of the current landscape: while leading LLMs perform reasonably well on smaller (128K) corpora, their performance degrades significantly as the scale approaches 1M tokens. More strikingly, standard RAG systems fail catastrophically on our benchmark, as their retrieval mechanisms are unable to gather the globally distributed evidence required. In contrast, memory-augmented agentic systems demonstrate substantially greater resilience, highlighting a promising direction for future research.

Our main contributions are:
\begin{enumerate}
    \item We introduce \textbf{CorpusQA}, the first large-scale benchmark for corpus-level analysis, featuring diverse domains, complex reasoning tasks, and a scale up to 10M tokens.
    \item We present a novel long-context data synthesis framework that programmatically generates complex questions with verifiable ground-truth answers. We further demonstrate that data synthesized by this framework can effectively enhance an LLM's long-context reasoning capabilities when used for training.
    \item Through comprehensive experiments, we demonstrate the severe limitations of current LLMs and RAG systems on this new task paradigm and provide strong evidence that memory-based architectures are a more promising path forward for true corpus-level reasoning.
\end{enumerate}

\section{Related Work}

\subsection{RAG Evaluation}
Retrieval-Augmented Generation (RAG) has become a standard paradigm for grounding LLMs in external knowledge. Consequently, a variety of benchmarks have been developed to evaluate RAG systems. For instance, \textbf{FRAMES}~\cite{krishna2024factfetchreasonunified} assesses multi-hop reasoning over structured and unstructured knowledge bases, while \textbf{CRAG}~\cite{yang2024crag} introduces a more dynamic evaluation of RAG pipelines, including self-correction and web search capabilities. However, these benchmarks are premised on a "sparse retrieval" assumption—that answers can be synthesized from a few retrieved document chunks. This paradigm fails for corpus-level analysis, where critical information is highly dispersed across the entire document set. Such tasks demand a global understanding, rendering sparse retrieval inherently insufficient.

\subsection{Long Context Evaluation}
Recent advancements have spurred the creation of benchmarks targeting long-context understanding. One category, including \textbf{LongBenchV2}~\cite{bai2024longbench} and \textbf{InfiniteBench}~\cite{zhang2024infty}, primarily evaluates a model's ability to recall facts or reason within a single, concatenated text. While they push the boundaries of context length, they do not address the challenge of integrating information scattered across a multitude of discrete documents.

Another category, multi-document QA benchmarks, takes a step closer to our paradigm. For example, \textbf{Loong}~\cite{wang2024leave} and \textbf{FanOutQA}~\cite{zhu2024fanoutqa} require reasoning across several documents. However, these benchmarks typically involve fewer than a dozen documents, and the total context size often remains within the capacity of modern long-context LLMs. In contrast, CorpusQA scales to hundreds of documents and millions of tokens, where evidence is so dispersed that simple retrieval or direct ingestion fails, necessitating more advanced, system-level architectures for global information synthesis.

% \subsection{Long Context Evaluation}

% Recent long-context benchmarks such as LongBenchV2~\cite{bai2024longbench} and InfiniteBench~\cite{zhang2024infty} extend inputs to the 100k–1M token range and evaluate multi-task reasoning within a single concatenated context. While they highlight scaling challenges, their design largely assumes that all necessary evidence can be ingested at once. In contrast, CorpusQA goes beyond by scaling to over 10M tokens and shifting the focus to corpus-level analysis, where evidence is scattered across hundreds of documents and answers require global aggregation and precise statistics.

% Some of the latest long-text benchmarks also emphasize information extraction and aggregation across multiple documents. For example, \cite{wang2024leave} propose Loong, a novel long-context benchmark requiring that all documents in a multi-document QA context be relevant for successful answering. FanOutQA~\cite{zhu2024fanoutqa} introduces a new benchmark targeting “fan-out” questions—a class of real-world multi-hop queries that demand reasoning across multiple documents and aggregating information about numerous entities. However, in these benchmarks, the evidence for each question is dispersed across at most four or five documents. Some powerful RAG systems can already handle such problems. In contrast, in corpus-level tasks, evidence may be scattered across hundreds of documents, making it nearly impossible for RAG systems to solve them. Tackling these problems requires either much stronger base LLMs or more advanced system-level solutions.

\section{The Corpus-level QA Benchmark}
\subsection{Overview}
\begin{figure*}[t]
  \centering
  \includegraphics[width=\textwidth]{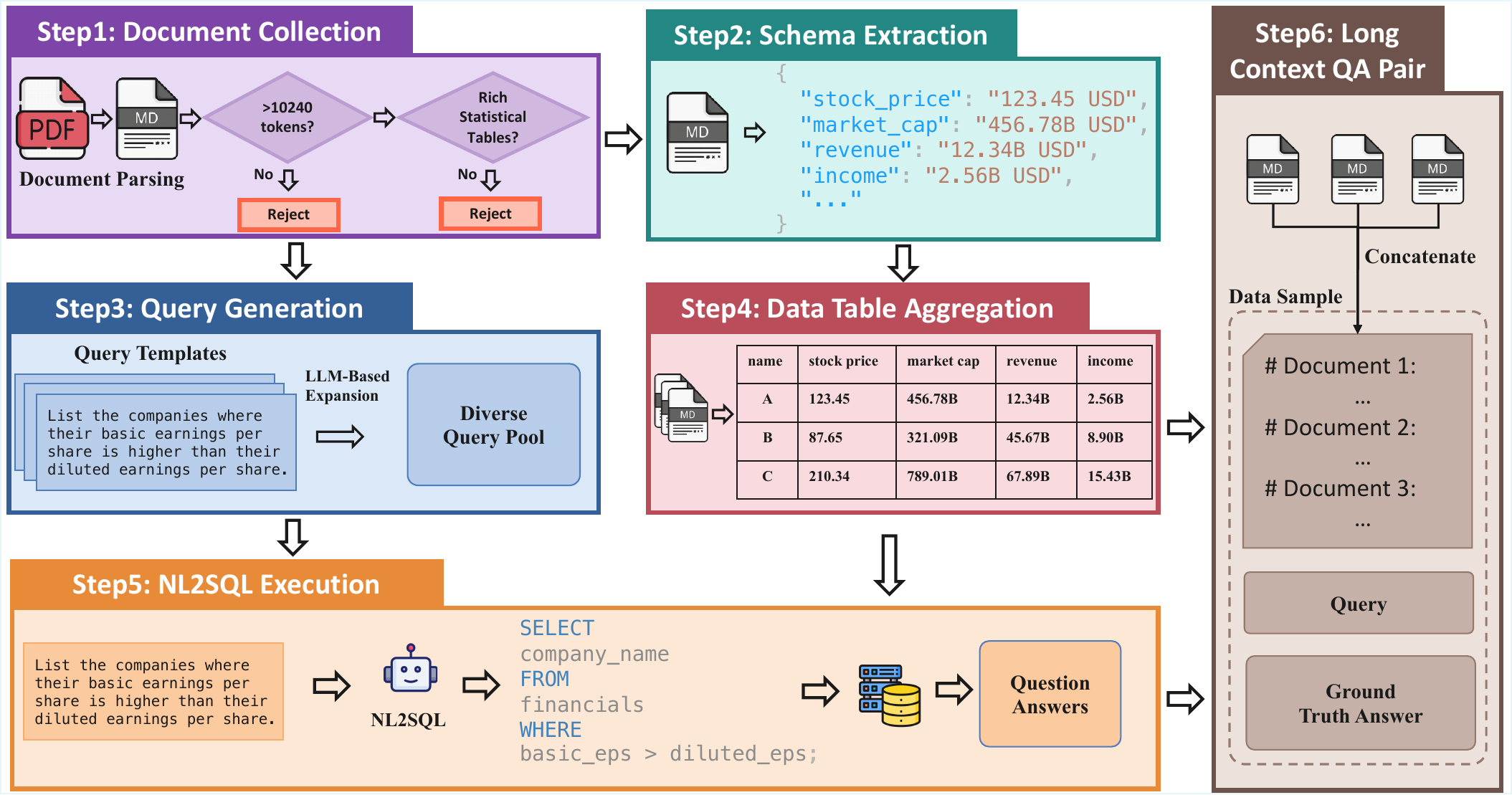}
  \caption{Overall data generation pipeline of CorpusQA. The process consists of six steps: (1) document collection and filtering to ensure long and data-rich inputs, (2) schema extraction for structured representation, (3) query generation via templates and LLM-based expansion, (4) data table aggregation across documents, (5) programmatic ground-truth generation using NL2SQL execution, and (6) final QA pair assembly by concatenating original documents with generated queries and answers. This design guarantees both realism and verifiable correctness for corpus-level reasoning tasks.}
  \label{fig:main}
\end{figure*}

Creating a benchmark for corpus-level analysis presents a significant challenge: how to generate complex, reasoning-intensive questions with verifiably correct answers when the task itself is designed to be difficult for even the most advanced LLMs. To overcome this, we designed a novel data generation pipeline that decouples the complex reasoning from the unstructured text. By first structuring the information, programmatically deriving ground-truth answers, and then re-presenting the original unstructured documents to the models being evaluated, we ensure 100\% factual accuracy and logical consistency. As illustrated in Figure \ref{fig:main}, our pipeline is composed of six main stages, which can be grouped into four key phases.

\subsection{Document Curation and Structuring}
The foundation of CorpusQA lies in high-quality, data-rich source documents. This phase involves carefully selecting appropriate documents and transforming their key information into a machine-readable format.

\paragraph{Step 1: Document Collection: } We first collect a large set of real-world PDF documents from diverse domains such as finance and academia. Each PDF is then parsed into Markdown to obtain a uniform, text-centric representation for downstream processing. To ensure the benchmark's focus on long-context and quantitative reasoning, we apply a two-stage filtering process. First, documents with fewer than 10,240 tokens are discarded. Second, we retain only those documents containing rich statistical tables and numerical data, as these are essential for constructing the complex analytical queries that define our benchmark.

\paragraph{Step 2: Schema Extraction} We perform schema extraction using a robust, multi-model voting approach to guarantee data quality. For each document, we query an ensemble of three state-of-the-art language models (Gemini 2.5 Pro~\cite{comanici2025gemini}, GPT-4.1~\cite{openai2025gpt4_1}, and Qwen-Max~\cite{alibaba2025qwen_max}) multiple times for each target field. A value is only considered valid if it represents a clear majority across all extraction results. This consensus-driven method filters out hallucinations and inconsistencies; documents that fail to produce a consensus for any field are discarded entirely. The resulting validated key-value pairs (e.g., stock\_price: "123.45 USD", revenue: "12.34B USD") are then structured into a consistent JSON-based schema for programmatic use in subsequent steps.

\subsection{LLM-Augmented Query Generation}
To ensure a diverse and challenging set of questions, we generate queries through a semi-automated process that combines human expertise with LLM capabilities.
\paragraph{Step 3: Query Generation} We first manually author a set of high-quality query templates that encapsulate various reasoning patterns, including comparison, ranking, conditional filtering, and multi-step calculations. An example template is: "List the companies where their basic earnings per share is higher than their diluted earnings per share." To increase linguistic diversity and prevent models from overfitting to specific phrasings, we leverage an LLM to paraphrase and expand these templates. This process results in a large and varied pool of natural language queries that are semantically grounded in the initial templates and categorized by difficulty (Easy, Medium, Hard) based on their logical and computational complexity. For a detailed breakdown of these difficulty levels with examples, please see Appendix \ref{app:difficulty_levels}.

% 在 wraptable 之前或内部设置 columnsep，控制文字与表格的距离

\begin{wraptable}{r}{0.4\textwidth} % <--- 关键修改：继续减小宽度至 0.31 (约31%)
    \vspace{-15pt}
    \centering
    \small
    \setlength{\tabcolsep}{4.5pt} % <--- 保持极小的列间距
    \renewcommand{\arraystretch}{1.1}
    
    \begin{tabular}{l c c r}
        \toprule
        \textbf{Category} & \textbf{Avg T.} & \textbf{Lang.} & \textbf{\#Inst.} \\
        \midrule
        \multicolumn{4}{c}{\textit{Domain}} \\
        \midrule
        Financial & 2.71M & ZH, EN & 664 \\
        Education & 3.12M & EN & 328 \\
        Real Est. & 3.59M & EN & 324 \\
        \midrule
        \multicolumn{4}{c}{\textit{Length Set}} \\
        \midrule
        Set1 (128K) & 92.5K & EN, ZH & 329 \\
        Set2 (1M)   & 905K  & EN, ZH & 329 \\
        Set3 (4M)   & 3.71M & EN, ZH & 329 \\
        Set4 (10M)  & 7.84M & EN, ZH & 329 \\
        \midrule
        \multicolumn{4}{c}{\textit{Difficulty}} \\
        \midrule
        Easy   & 3.11M & EN, ZH & 296 \\
        Medium & 3.11M & EN, ZH & 648 \\
        Hard   & 3.20M & EN, ZH & 372 \\
        \bottomrule
    \end{tabular}
    \caption{Statistics of the dataset.}
    \label{tab:dataset_breakdown}
    \vspace{-40pt}
\end{wraptable}
\subsection{Programmatic Ground Truth Generation}
This phase is the cornerstone of our methodology, guaranteeing the correctness of our benchmark's answers without relying on a fallible LLM's reasoning over long text.

\paragraph{Step 4: Data Table Aggregation} The structured JSON objects extracted from every document in a given corpus are aggregated to construct a single, global data table. In this table, each row typically represents an entity (e.g., a company, a university), and each column corresponds to a specific attribute from our schema. This table serves as the canonical, structured knowledge base for the entire document set.

\paragraph{Step 5: NL2SQL Execution} To generate a verifiable answer for each query, we adopt a robust programmatic approach. A natural language query from our pool is first translated into an executable SQL statement using a state-of-the-art LLM. For instance, the example query from Step 3 would be converted to SELECT company\_name FROM financials WHERE basic\_eps > diluted\_eps;. This SQL statement is then executed directly against the aggregated data table from Step 4. This method ensures that every ground truth answer is 100\% accurate and logically derived from the complete source data, establishing an unimpeachable standard for evaluation.

\subsection{Final QA Pair Assembly}
In the final stage, we assemble the components into the complete benchmark instance that will be presented to the models.

\paragraph{Step 6: Long Context QA Pair:} The original, full-text Markdown documents selected in Step 1 are concatenated in their entirety to form the final long-context input. Each instance in CorpusQA is thus a triplet comprising: (1) the concatenated document corpus as the context, (2) a natural language query from our generated pool, and (3) the programmatically derived ground truth answer. This final structure challenges a model to perform the sophisticated, corpus-level reasoning over raw, unstructured text that we have systematically executed on a structured backend.

\subsection{Data Quality Verification}
To validate the integrity of our data generation pipeline, we conducted a rigorous quality check on the schema extraction stage, which forms the basis for our programmatic ground-truth generation. For the education\_en and financial\_zh domains, which rely on model-based extraction from unstructured text, we randomly sampled 1,000 extracted schemas per domain for manual verification. The financial\_en and real\_estate\_en domains were exempt, as their data was derived directly from structured sources, ensuring its fidelity. The initial consistency between the model's extractions and human verification was very high: 94.13\% for education\_en and 98.97\% for financial\_zh. Critically, a secondary review of all inconsistent cases by different human annotators revealed that the LLM's original extraction was correct in every instance of disagreement. This finding underscores the robustness of our multi-model voting approach and confirms the near-perfect accuracy of the structured data used to generate the benchmark's ground-truth answers.

\section{Experiments}
\begin{table*}[t]
\centering
\setlength{\tabcolsep}{5pt}
\renewcommand{\arraystretch}{1.15}

\resizebox{\textwidth}{!}{
    \begin{tabular}{lccccc} 
    \toprule
    \textbf{Model} & \textbf{Financial-zh} & \textbf{Financial-en} & \textbf{Education-en} & \textbf{Real Estate-en} & \textbf{Overall} \\
    \midrule
    
    % --- Set 1 区域 ---
    \multicolumn{6}{c}{\textbf{Set1 (128K)}} \\
    \midrule % 【修改点1】添加横线，将子标题和下方数据隔开，压实结构
    \rowcolor{gray!10}
    Gemini-2.5-Pro             & 89.16 & 80.72 & 76.83 & 74.07 & 80.19 \\
    Gemini-2.5-Flash           & 83.13 & 74.70 & 74.39 & 79.01 & 77.81 \\
    \rowcolor{gray!10}
    GPT-5                      & 93.98 & 78.31 & 81.71 & 74.07 & 82.02 \\
    GPT-5-Mini                 & 93.98 & 78.31 & 74.39 & 72.84 & 79.88 \\
    \rowcolor{gray!10}
    Qwen3-235B-A22B-Thinking & 90.36 & 74.70 & 70.73 & 76.54 & 78.08 \\
    Qwen3-30B-A3B-Thinking     & 83.13 & 61.45 & 64.63 & 69.14 & 69.59 \\
    \rowcolor{gray!10}
    DeepSeek-R1-0528           & 83.13 & 73.49 & 78.05 & 64.20 & 74.72 \\
    MiniMax-M1-80k             & 73.49 & 67.07 & 76.83 & 66.67 & 71.02 \\
    
    % --- Set 2 区域 ---
    \midrule % 【修改点2】添加横线，彻底分割 Set1 和 Set2，消除“悬浮感”
    \multicolumn{6}{c}{\textbf{Set2 (1M)}} \\
    \midrule % 【修改点3】同上，子标题下划线
    \rowcolor{gray!10}
    Gemini-2.5-Pro             & 50.60 & 48.19 & 65.85 & 38.27 & 50.73 \\
    Gemini-2.5-Flash           & 26.25 & 43.37 & 48.78 & 30.86 & 37.31 \\
    \rowcolor{gray!10}
    MiniMax-M1-80k             & 12.05 & 9.64  & 7.32  & 17.50 & 11.63 \\
    
    \bottomrule
    \end{tabular}
}
\caption{Accuracy (\%) of base LLMs across four domains under two context-length settings. 
Set1 (128K) corresponds to the maximum context length supported by most frontier models, 
while Set2 (1M) includes the few base LLMs that claim million-token support. 
Results show strong performance at 128K but substantial degradation at 1M.}
\label{tab:base-llm-results}
\end{table*}

\begin{table*}[t]
\centering
\setlength{\tabcolsep}{5pt}
\renewcommand{\arraystretch}{1.15}

\resizebox{\textwidth}{!}{
    \begin{tabular}{lccccc}
    \toprule
    \rowcolor{white}
    \textbf{Model} & \textbf{Financial-zh} & \textbf{Financial-en} & \textbf{Education-en} & \textbf{Real Estate-en} & \textbf{Overall} \\
    \midrule
    
    % --- Set 1 ---
    \multicolumn{6}{c}{\textbf{Set1 (128K)}} \\
    \midrule % 【新增】分隔标题与数据
    \rowcolor{gray!10}
    Memory Agent - Gemini-2.5-Pro & 79.01 & 57.58 & 65.82 & 67.90 & 67.58 \\
    RAG - Gemini-2.5-Pro          & 71.08 & 51.81 & 58.54 & 41.98 & 55.85 \\
    
    % --- Set 2 ---
    \midrule % 【修改】用线分隔板块
    \multicolumn{6}{c}{\textbf{Set2 (1M)}} \\
    \midrule % 【新增】分隔标题与数据
    \rowcolor{gray!10}
    Memory Agent - Gemini-2.5-Pro & 56.79 & 45.78 & 41.46 & 35.80 & 44.95 \\
    RAG - Gemini-2.5-Pro          & 10.39 & 18.07 & 25.00 & 16.05 & 17.38 \\
    
    % --- Set 3 ---
    \midrule % 【修改】用线分隔板块
    \multicolumn{6}{c}{\textbf{Set3 (4M)}} \\
    \midrule % 【新增】分隔标题与数据
    \rowcolor{gray!10}
    Memory Agent - Gemini-2.5-Pro & 19.28 & 12.05 & 32.93 & 28.12 & 22.14 \\
    RAG - Gemini-2.5-Pro          & $1.20^{\dagger}$ & 3.61 & $3.70^{\dagger}$ & $1.22^{\dagger}$ & 2.43 \\
    
    % --- Set 4 ---
    \midrule % 【修改】用线分隔板块
    \multicolumn{6}{c}{\textbf{Set4 (10M)}} \\
    \midrule % 【新增】分隔标题与数据
    \rowcolor{gray!10}
    Memory Agent - Gemini-2.5-Pro & 7.89 & 9.64 & 14.63 & 12.35 & 11.13 \\
    RAG - Gemini-2.5-Pro          & $3.61^{\dagger}$ & $1.20^{\dagger}$ & $0.00^{\dagger}$ & $0.00^{\dagger}$ & 1.20 \\
    
    \bottomrule
    \end{tabular}
}
\caption{Accuracy (\%) of \texttt{Memory Agent} and \texttt{RAG}, both built on \texttt{Gemini-2.5-Pro}, across four domains under four different corpus scales (128K–10M tokens). Higher values indicate better performance. $^{\dagger}$ denotes cases where the number of retrieved chunks by RAG was smaller than the number of documents, making accurate answers impossible.}
\label{tab:memory-rag-results}
\end{table*}

\subsection{Experimental Setup}
\paragraph{Evaluation Targets}
Our benchmark is divided into four context lengths: 128k, 1M, 4M, and 10M. 
Among them, the 128k and 1M settings are mainly used for directly testing the long-context capabilities of LLMs, while the 4M and 10M settings are primarily used to evaluate the effectiveness of advanced strategies such as retrieval-augmented generation and memory agents. For long-context evaluation, we consider both closed-source and open-source models. 
Closed-source models include \texttt{Gemini-2.5-Pro}~\cite{comanici2025gemini}, \texttt{Gemini-2.5-Flash}~\cite{comanici2025gemini}, \texttt{GPT-5-2025-08-07}~\cite{openai2025gpt5}, \texttt{GPT-5-Mini-2025-08-07}~\cite{openai2025gpt5}. 
Open-source models include the \texttt{Qwen3} family~\cite{yang2025qwen3} (\texttt{Qwen3-235B-Thinking-2507} and \texttt{Qwen3-30B-Thinking-2507}, \texttt{DeepSeek-R1-0528}~\cite{guo2025deepseek}, and \texttt{Minimax-M1-80k}~\cite{chen2025minimax}. 
All of the above models are evaluated on the 128k setting, while the 1M setting is tested with \texttt{Gemini-2.5-Pro}, \texttt{Gemini-2.5-Flash}, and \texttt{Minimax-M1} due to their extended context window support. 

For the 4M and 10M settings, where the total corpus length significantly exceeds any available model’s context window, we focus on system-level solutions:  

\textbf{Retrieval-Augmented Generation.}  
RAG systems operate by first segmenting the corpus into retrievable chunks, then retrieving a small subset relevant to the query, and finally generating answers conditioned on both the retrieved evidence and the query. This approach reduces the effective input size but relies heavily on retrieval recall and chunk-level information density. Moreover, because corpus-level tasks often involve evidence scattered across the entire context, a large number of chunks must be retrieved to ensure sufficient coverage and accuracy of the final answer.

\textbf{Memory Agents.}  
The Memory Agent~\cite{yu2025memagent} framework is designed for ultra-long text processing. It operates by processing a document in chunks, iteratively updating a fixed-size memory buffer. The final output is generated solely from the information consolidated in this memory. We evaluate two implementations of this approach: (1) \texttt{MemAgent-14B}, a specialist model fine-tuned for this task by~\cite{yu2025memagent}, and (2) general-purpose LLM (\texttt{Gemini-2.5-Pro}) instructed to follow the same workflow without task-specific tuning. 

\paragraph{Evaluation Metrics} 
Many answers in our benchmark involve numbers, percentages, or named entities that can appear in different but equivalent forms.  
This makes exact string matching unreliable for correctness evaluation.  

To address this, we use an LLM-as-a-Judge~\cite{zheng2023judging,gu2024survey} approach: a strong LLM is prompted to assess whether a model’s prediction is semantically equivalent to the reference answer.  
We then compute accuracy as the percentage of predictions judged correct.  
This method provides a more flexible and robust evaluation than strict matching, especially for reasoning-heavy and numerically precise tasks.  
We use \texttt{DeepSeek-V3}\cite{liu2024deepseek} as the judge model, and the prompt template is provided in the Appendix. 

\paragraph{Implement Details} We set temperature = 0 to eliminate randomness and keep other hyper-parameters default. For API-Based LLMs, we directly utilize the official API for testing. As for open-source models, we conduct experiments on a server with 8×A100 80GB.

\subsection{Main Results}
Most base LLMs currently support context lengths of up to 128k tokens, and we evaluate ten recent models under this setting. In addition, a few base LLMs are able to process 1M-token contexts, for which we evaluate three models. The results across both the 128k and 1M settings are reported in Table \ref{tab:base-llm-results}. For larger-scale settings (4M and 10M tokens), which exceed the context limits of base LLMs, we rely on system-level methods—retrieval-augmented generation and Memory Agents to handle ultra-long contexts. Their performance is evaluated across the 128k to 10M settings, with results summarized in Table \ref{tab:memory-rag-results}. 

Table \ref{tab:base-llm-results} shows that closed-source models achieve the best results at 128K, with \texttt{GPT-5} (82.02) and \texttt{Gemini-2.5-Pro} (80.19) leading the benchmark. When extended to 1M tokens, performance drops substantially for all models, yet \texttt{Gemini-2.5-Pro} remains comparatively strong at 50.73, outperforming other baselines by a large margin. Among open-source models, \texttt{Qwen3-235B-Thinking} (78.08) and \texttt{DeepSeek-R1} (74.72) perform competitively at 128K, but instruction-tuned variants and smaller models degrade severely, with \texttt{MiniMax-M1-80k} collapsing to 11.63 at 1M. Overall, while both closed- and open-source LLMs handle 128K contexts reasonably well, only top-tier models such as \texttt{Gemini-2.5-Pro} show partial resilience at the million-token scale. 

% 确保在导言区添加了 \usepackage{caption}

\begin{wrapfigure}{R}{0.55\textwidth} % 宽度设为 0.55 以容纳表格宽度
    \centering
    \vspace{-15pt}
    
    % --------- 上半部分：图片 ---------
    \includegraphics[width=\linewidth]{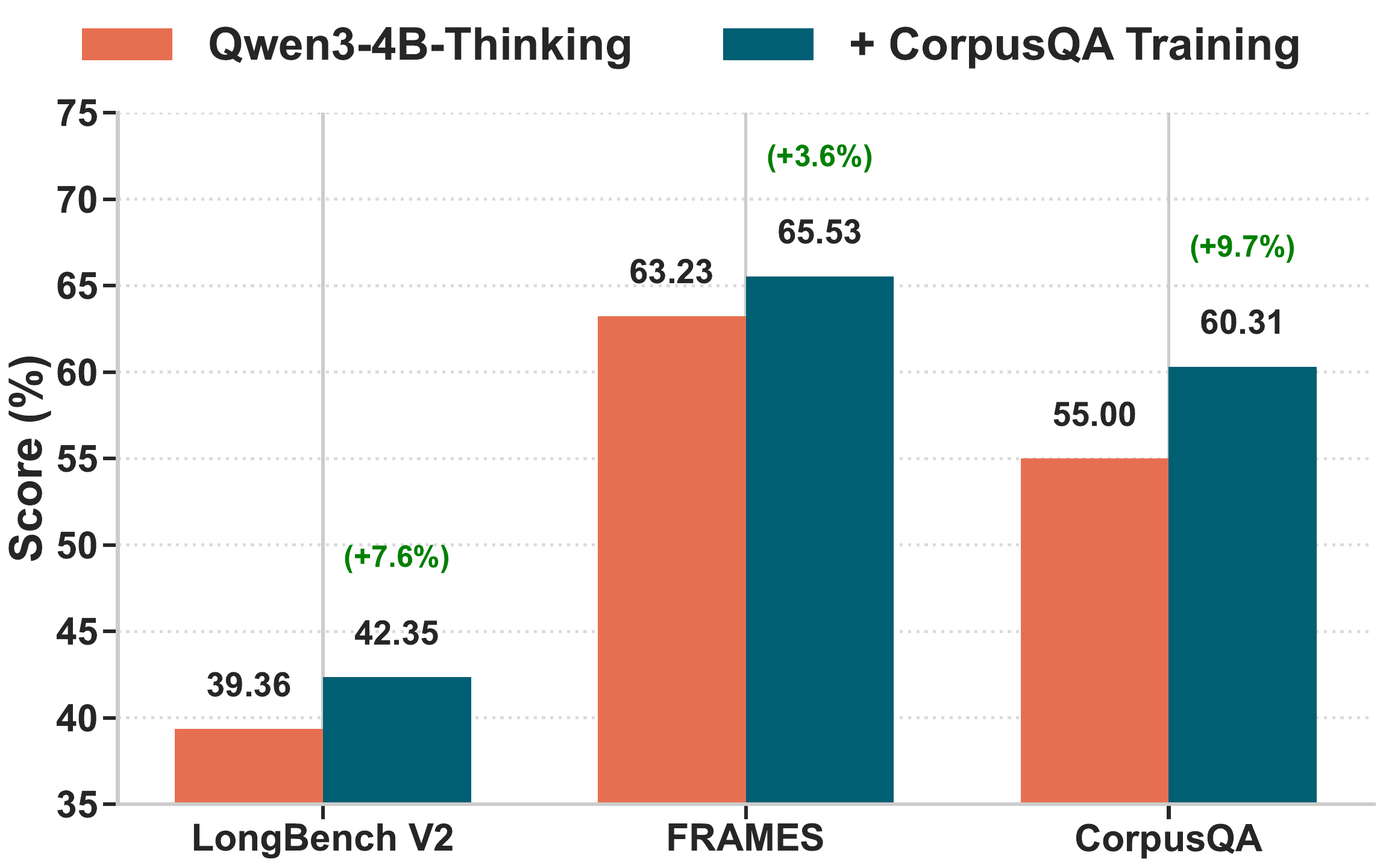}
    \caption{Impact of fine-tuning with synthesized data. Training a model on data generated by our synthesis framework not only improves in-domain performance (CorpusQA) but also generalizes to enhance performance on external long-context benchmarks (LongBenchV2 and FRAMES).}
    \label{fig:finetuning-results}
    
    \vspace{20pt} % 图片和表格之间的间距
    
    % --------- 下半部分：表格 ---------
    \footnotesize 
    \renewcommand{\arraystretch}{1.0} 
    \setlength{\tabcolsep}{3pt}       
    \begin{tabular}{l l c c c}
        \toprule
        \textbf{Context} & \textbf{Method} & \textbf{Acc.(\%)} & \makecell[c]{\textbf{Time}\\\textbf{(min)}} & \makecell[c]{\textbf{Timeout}\\\textbf{(\%)}} \\
        \midrule
        % --- 128K Section ---
        \multirowcell{3}{\textbf{128K}}
        & Human        & 75.0 & 17.8 & 6.6 \\
        & Base LLM     & 66.7 & \textbf{2.1}  & 0 \\
        & Memory Agent & 50.0 & 1.4  & 0 \\
        \midrule
        % --- 1M Section ---
        \multirowcell{3}{\textbf{1M}}
        & Human        & 50.0 & 31.9 & 33.3 \\
        & Base LLM     & 20.0 & 120.7 & 0 \\
        & Memory Agent & 33.3 & 69.7  & 0 \\
        \midrule
        % --- 4M Section ---
        \multirowcell{3}{\textbf{4M}}
        & Human        & 0.0  & --    & 100.0 \\
        & Base LLM     & \multicolumn{3}{c}{\textit{Context Limit Exceeded}} \\
        & Memory Agent & 25.0 & 323.5 & 0 \\
        \midrule
        % --- 10M Section ---
        \multirowcell{3}{\textbf{10M}}
        & Human        & 0.0  & --    & 100.0 \\
        & Base LLM     & \multicolumn{3}{c}{\textit{Context Limit Exceeded}} \\
        & Memory Agent & 16.7 & 789.3 & 0 \\
        \bottomrule
    \end{tabular}
    
    % 关键点：在 wrapfigure 环境里给表格加标题，必须用 \captionof{table}
    \captionof{table}{
        \textbf{Human on CorpusQA.} Comparison of a human expert, a base LLM, and our agent. 
        \textbf{Acc.}: Accuracy. \textbf{Time}: Average time in minutes. \textbf{Timeout}: Percentage of tasks where the subject timed out ($>$90min) or gave up.
        Base LLMs cannot process 4M/10M contexts due to architectural limits.
    }
    \label{tab:human_vs_ai}
    \vspace{-47pt}
\end{wrapfigure}

\subsection{The Collapse of RAG and the Promise of Memory Agents}

Table \ref{tab:memory-rag-results} compares \texttt{Memory Agent} and \texttt{RAG}, both built on \texttt{Gemini-2.5-Pro}, from 128K to 10M tokens. At 128K, \texttt{Memory Agent} already outperforms \texttt{RAG} (67.58 vs.\ 55.85). As scale increases, the gap widens: at 1M tokens, \texttt{Memory Agent} retains moderate performance (44.95), while \texttt{RAG} drops sharply (17.38); at 4M and 10M tokens, \texttt{RAG} nearly fails, whereas \texttt{Memory Agent} remains functional. 

These results reveal a core challenge of corpus-level analysis: \textbf{high information dispersion}. Retrieval-only methods cannot capture evidence scattered across thousands of documents, leading to collapse under ultra-long contexts. In contrast, memory mechanisms provide more robust aggregation, though still far from solving the task. By scaling to 10M tokens, our benchmark not only reflects the growing context windows of base LLMs but also offers a platform to evaluate advanced system-level methods such as memory, compression, and agentic approaches.

\subsection{Impact of Corpus-Level Training}
A key advantage of our data synthesis framework is its ability to generate not only evaluation benchmarks but also high-quality training data. To demonstrate this, we conducted a pilot experiment to assess the impact of training on data synthesized by our framework. We synthesized a new training set of 480 samples and used it to fine-tune the open-source \texttt{Qwen3-4B-Thinking} model for 25 steps with Group Relative Policy Optimization (GRPO)~\cite{shao2024deepseekmath}. We then evaluated the model's performance on our CorpusQA test set (in-domain) and on two external long-context benchmarks, LongBenchV2 and FRAMES, to measure out-of-domain generalization.

The results, presented in Figure \ref{fig:finetuning-results}, are promising. The fine-tuned model achieved a significant gain on CorpusQA, demonstrating its ability to learn the corpus-level analysis task. More interestingly, the model also showed notable improvements on both LongBenchV2 and FRAMES. This suggests that training on the complex, globally-aware reasoning tasks generated by our framework can enhance a model's broader long-context capabilities.

\subsection{Human Performance}
To contextualize the benchmark's difficulty, we conducted a comparative study between a human expert and two AI systems: a base LLM (\texttt{Qwen3-30B-A3B-Thinking}) and a Memory Agent built upon the same model. We sampled 12 question-answer pairs at each scale (128K, 1M, 4M, and 10M tokens). To prevent humans from spending too much time on a single problem, we set a time limit of 90 minutes, after which they give up. The results are summarized in Table \ref{tab:human_vs_ai}.

The findings reveal a clear performance hierarchy that shifts dramatically with scale. At 128K tokens, the task is challenging but manageable: the human expert is most accurate (75.0\%) but slow, while the base LLM offers a strong balance of speed and accuracy (66.7\%). However, the 1M token mark represents a critical tipping point where human and base LLM performance plummets, and beyond this, the task becomes demonstrably superhuman. At 4M and 10M tokens, the human expert failed every task and the base LLM is architecturally incapable of processing the input, leaving the Memory Agent as the only viable approach, which remained operational and achieved non-trivial accuracy (25.0\% at 4M and 16.7\% at 10M). This experiment powerfully underscores two conclusions: first, CorpusQA presents a challenge that rapidly exceeds human cognitive limits, validating its difficulty; second, it provides compelling evidence that for truly large-scale corpus analysis, simply expanding an LLM's context window is an inefficient strategy. Scalable, resilient architectures like the Memory Agent are not just an alternative but a necessity to tackle reasoning at this scale.

\section{Conclustion}
In this work, we introduced CorpusQA, a new benchmark scaling to 10 million tokens designed to evaluate corpus-level reasoning, a task defined by high evidence dispersion where traditional benchmarks fall short. Our experiments reveal that while even top-tier LLMs struggle at scale, standard RAG systems collapse entirely, proving inadequate for holistic analysis. In contrast, memory-augmented agentic architectures demonstrate far greater resilience, suggesting a promising direction for future research. Our findings underscore a critical need to shift focus from merely extending context windows to developing novel architectures capable of global information synthesis, a challenge for which CorpusQA now provides a crucial yardstick.

\section*{Limitations} 
This work has two primary limitations. First, our evaluation relies on an LLM-as-a-Judge, which is less deterministic than a fully rule-based system. Second, our experiments did not extend to more complex agentic architectures, such as those performing deep research~\cite{li2025websailor,li2025websailorv2,tongyidr}, leaving their performance on CorpusQA as an important area for future investigation.

\bibliographystyle{iclr2025_conference}
\bibliography{iclr2025_conference}

\appendix
\section{Reasoning Models vs. Instruction Models}
\begin{figure}[th]
  \centering
  \includegraphics[width=0.75\columnwidth]{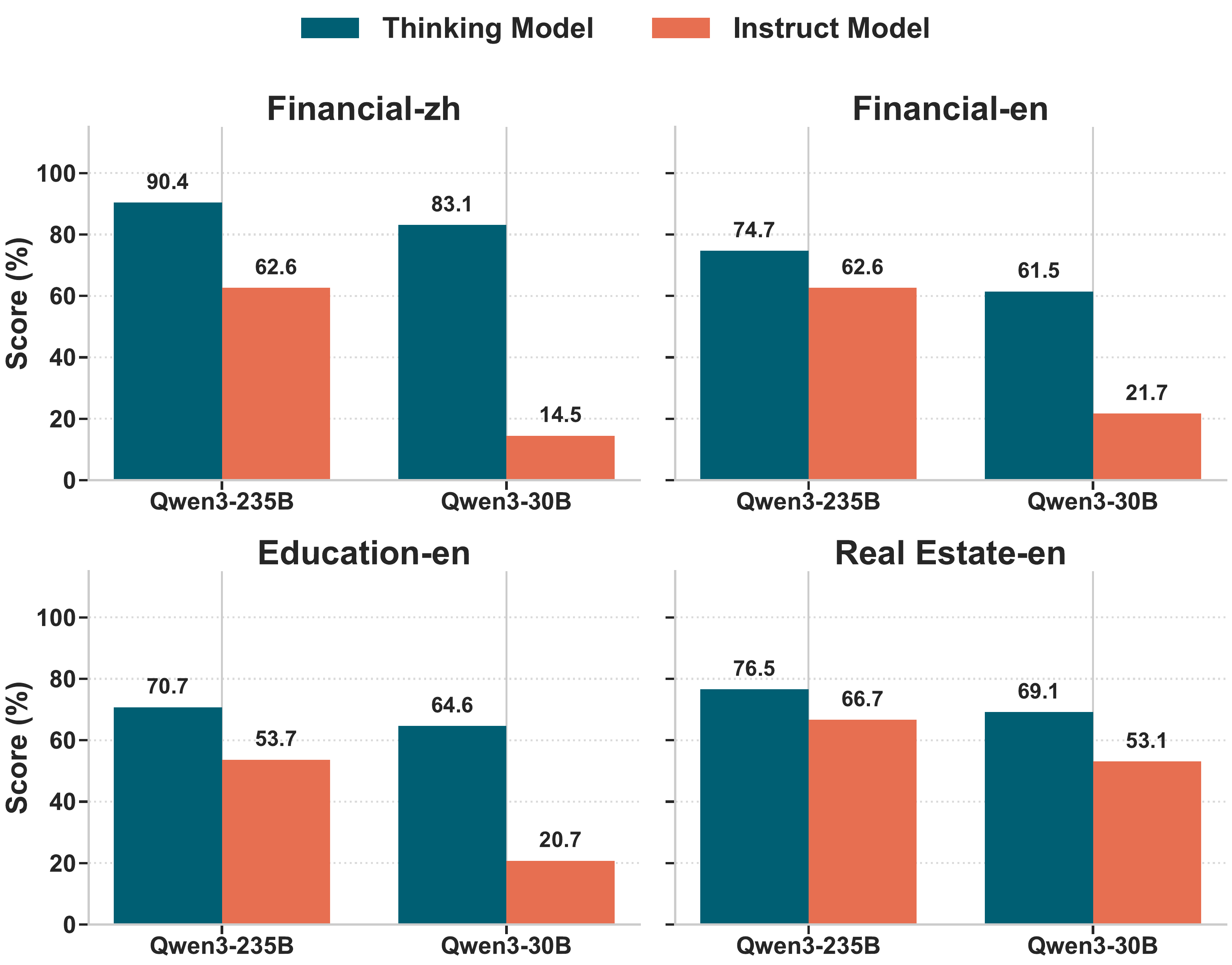}
  \caption{Comparison of Thinking vs. Instruct models on the 128K setting. Reasoning-oriented models (Thinking) consistently outperform their instruction-tuned counterparts, highlighting the benchmark's sensitivity to complex reasoning over simple instruction following.}
  \label{fig:thinking_comparison}
\end{figure}

Our benchmark specifically targets long-context reasoning, and the results in Figure \ref{fig:thinking_comparison} clearly show that reasoning-oriented models substantially outperform their instruction-tuned counterparts (e.g., \texttt{Qwen3-235B-Thinking}: 78.08 vs. \texttt{Qwen3-235B-Instruct}: 61.41). This consistent gap highlights a unique feature of our benchmark: it is sensitive to reasoning ability rather than alignment alone, making it particularly effective at revealing the trade-off introduced by instruction tuning in long-text scenarios.

\section{Query Difficulty Levels}
\label{app:difficulty_levels}

The queries in CorpusQA are classified into three difficulty levels to systematically evaluate different aspects of corpus-level reasoning. Below are the definitions and representative examples for each category.

\paragraph{Easy: Basic Aggregation and Single-Dimension Analysis}
This level tests for simple data retrieval, filtering, and sorting.
\begin{itemize}
    \item \textbf{Simple Aggregation:} e.g., "List the top 5 universities with the highest total enrollment."
    \item \textbf{Single-Condition Filtering:} e.g., "Find universities with more than 1,000 part-time faculty members."
    \item \textbf{Basic Ranking:} e.g., "Rank the top 10 universities by the number of transfer students."
\end{itemize}

\paragraph{Medium: Numerical Calculation and Multi-Condition Analysis}
This level requires combining multiple criteria and performing basic arithmetic operations.
\begin{itemize}
    \item \textbf{Ratio Calculation:} e.g., "Calculate the admission rate for each university."
    \item \textbf{Multi-Condition Combination:} e.g., "Which universities have a graduation rate above 90\% and fewer than 5,000 undergraduate students?"
    \item \textbf{Simple Custom Metrics:} e.g., "What is the student-to-faculty ratio for each institution?"
    \item \textbf{Grouped Comparison:} e.g., "Compare the average graduation rates of large versus small universities."
\end{itemize}

\paragraph{Hard: Custom Metrics and Multi-Step Reasoning}
This level involves complex, multi-step calculations, custom-defined business logic, and advanced analytical reasoning.
\begin{itemize}
    \item \textbf{Multi-Step Calculation:} e.g., "For all universities with a first-year admission rate below 50\%, calculate the average six-year graduation rate. Then, find the universities in this group whose individual graduation rate is more than 15 percentage points above this calculated average."
    \item \textbf{Complex Business Logic:} e.g., "Which companies in Q1 2025 had a positive 'operating safety margin' (defined as Net cash from operations - Cash paid to employees - Financial expenses), where this margin also exceeded their net profit attributable to shareholders?"
    \item \textbf{Advanced Metric Construction:} e.g., "Identify companies with high financial risk, defined by the criterion: $(X_1 \times 1.2 + X_2 \times 1.4 + X_3 \times 3.3) < 1.8$, where $X_1 = (\text{Current Assets} - \text{Accounts Payable} - \text{Contract Liabilities}) / \text{Total Assets}$, $X_2 = \text{Undistributed Profit} / \text{Total Assets}$, and $X_3 = \text{Operating Revenue} / \text{Total Assets}$."
\end{itemize}

\section{Data Template}
To ensure consistent and fair evaluation, all models were tested using a standardized prompt structure. The template below details the format, which includes the concatenated corpus of documents, the specific question, and a strict set of instructions on output formatting.

\begin{promptbox}

% 使用 tabular 环境来替代 description，以实现对齐且无缩进
\begin{description}
    % --- 修改在这里 ---
    \item[\# Document 1:] \verb|<document1>|
    \item[\# Document 2:] \verb|<document2>|
    \item[\# Document 3:] \verb|<document3>|
    \item[...]
    % --- 修改在这里 ---
    \item[\# Question:] \verb|<question>|
\end{description}

\vspace{0.5em}
\textbf{\# Output requirements:}

% 使用 leftmargin=0pt 移除 enumerate 列表的缩进
\begin{enumerate}[leftmargin=1.5em, labelindent=0em, itemsep=0.5ex, topsep=0.5ex]
    \item Your answer can only be one of the following types: a list of strings (including empty list "[]"), a single short string, a number (floating point numbers should retain 2 decimal places, without units), or a percentage (retain 2 decimal places).
    \item To output the date, use the format "YYYY-MM".
    \item If multiple files contain a conflicting value for the same item, please use the value from the latest file.
    \item Please place the final answer at the end, and strictly use the format: \texttt{The answer is: xxx}.
\end{enumerate}

\vspace{0.5em}
\textbf{\# Answer format examples:}

% 使用 leftmargin=0pt 移除 itemize 列表的缩进
\begin{itemize}[leftmargin=1.5em, labelindent=0em, itemsep=0.5ex, topsep=0.5ex]
    \item For a list of strings: \texttt{The answer is: ["NETFLIX INC", "H\&R BLOCK INC"]}
    \item For a single string: \texttt{The answer is: "South"}
    \item For a number: \texttt{The answer is: 1234567.89}
    \item For a percentage: \texttt{The answer is: 12.34\%}
\end{itemize}

\end{promptbox}

\section{LLM-as-a-Judge Prompt}
Given that answers can be numerically complex or expressed in varied but equivalent phrasings, simple string matching is insufficient for accurate evaluation. We therefore employ an LLM-as-a-Judge approach. The following prompt instructs the judge model to assess the semantic equivalence between a model's generated answer and the ground truth, ensuring a robust and fair scoring process.

\begin{promptbox}
You are an expert in verifying if two answers are the same. Your input is a problem and two answers, Answer 1 and Answer 2. Your task is to determine if they are equivalent, without attempting to solve the original problem.
\par
Compare the answers to verify they represent identical values or meaning, even when written in different forms or notations.

\vspace{1em}
Your output \textbf{must} follow the following format:
\begin{enumerate}
    \item Provide an explanation for why the answers are equivalent or not.
    \item Then provide your final answer in the form of: [[YES]] or [[NO]]
\end{enumerate}

\vspace{1em} % 在列表后增加一点间距
\textbf{Problem:} \verb|<question>| \\
\textbf{Answer 1:} \verb|<model_answer>| \\
\textbf{Answer 2:} \verb|<ground_truth>|
\end{promptbox}

\end{document}